\documentclass[conference]{IEEEtran}
\IEEEoverridecommandlockouts
\usepackage{cite}
\usepackage{amsmath,amssymb,amsfonts}
\usepackage{algorithmic}
\usepackage{hyperref}
\usepackage{subcaption}
\usepackage{multirow}

\usepackage{graphicx}  
\usepackage{textcomp}
\usepackage{xcolor}
\def\BibTeX{{\rm B\kern-.05em{\sc i\kern-.025em b}\kern-.08em
    T\kern-.1667em\lower.7ex\hbox{E}\kern-.125emX}}
\begin{document}

\title{Beyond Point Matching: Evaluating Multiscale Dubuc Distance for Time Series Similarity\\
\thanks{This work was supported by NSF awards 2209912, 2433781, and 2511630.}
}

\author{
    \IEEEauthorblockN{Azim~Ahmadzadeh\IEEEauthorrefmark{1},
    Mahsa~Khazaei, 
    Elaina~Rohlfing}
    \IEEEauthorblockA{
        Department of Computer Science,
        University of Missouri-St. Louis,
        MO, USA\\
        Email: \IEEEauthorrefmark{1}ahmadzadeh@umsl.edu \\
        \vspace{-1.1cm}
    }
}


\maketitle

\begin{abstract}
Time series are high-dimensional and complex data objects, making their efficient search and indexing a longstanding challenge in data mining. Building on a recently introduced similarity measure, namely Multiscale Dubuc Distance (MDD), this paper investigates its comparative strengths and limitations relative to the widely used Dynamic Time Warping (DTW). MDD is novel in two key ways: it evaluates time series similarity across multiple temporal scales and avoids point-to-point alignment. We demonstrate that in many scenarios where MDD outperforms DTW, the gains are substantial, and we provide a detailed analysis of the specific performance gaps it addresses. We provide simulations, in addition to the 95 datasets from the UCR archive, to test our hypotheses. Finally, we apply both methods to a challenging real-world classification task and show that MDD yields a significant improvement over DTW, underscoring its practical utility.
\end{abstract}


\section{Introduction}
    Time series, or more generally, ordered high-dimensional data types, have become increasingly prevalent with the rise of powerful computational tools and machine learning techniques. In this study, we adopt the term time series as an umbrella label for all such sequential data. A central challenge in analyzing time series lies in defining and measuring similarity. Similarity is inherently subjective, shaped by the specific goals and nuances of a given application. traces in handwritten notes, gestures in sign language, movements of robotic arms, trajectory of GPS-tracked objects, and measurements captured by sensors are all examples of applications in which temporal and/or spatial alignment is crucial. The existing literature has produced a rich landscape of similarity measures, each tailored to specific assumptions and use cases. Navigating this landscape often requires expertise to select or adapt the appropriate tool for the task at hand.

    \begin{figure}[t]
        \centering
        \includegraphics[width=1\linewidth]{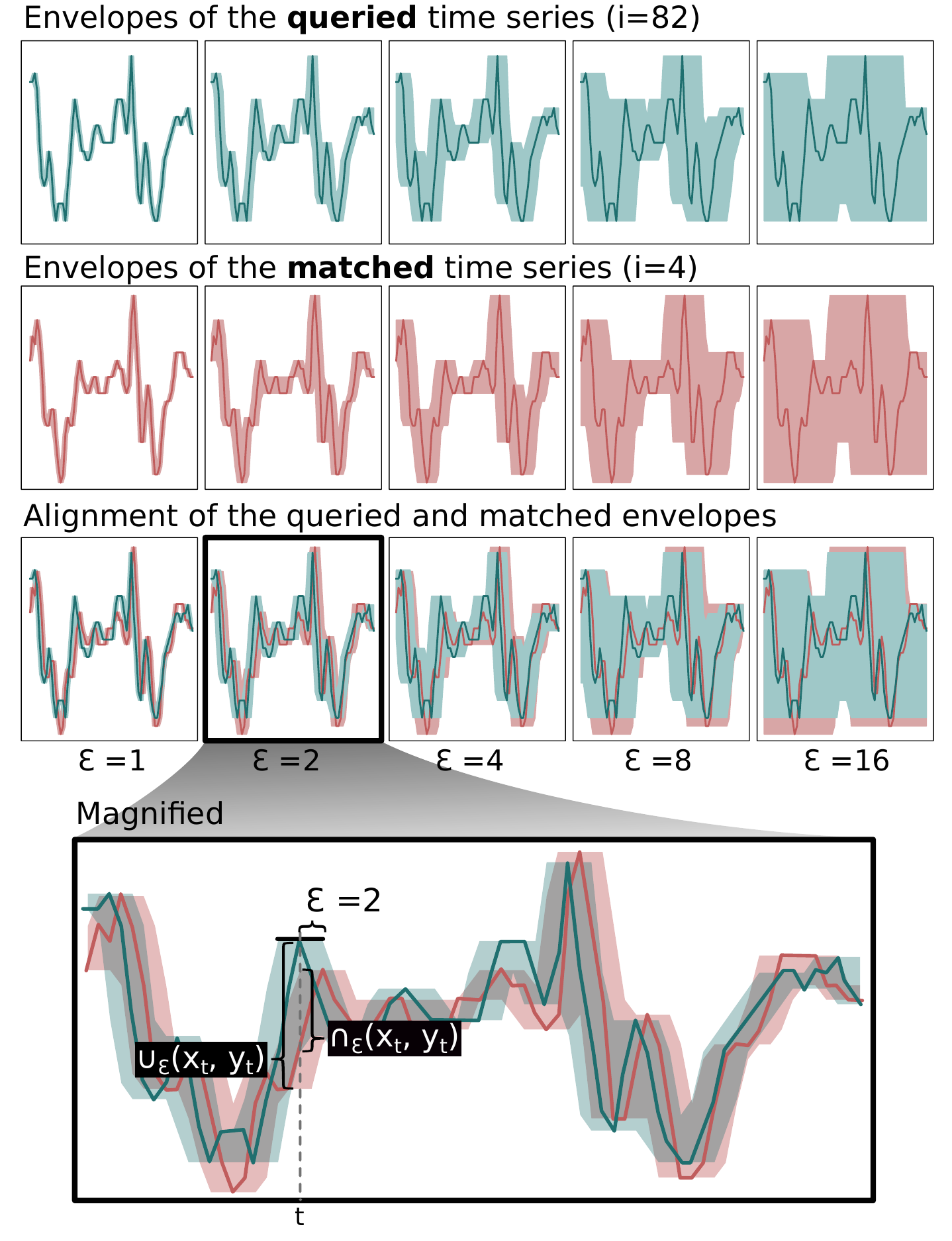}
        \caption{The illustration shows how MDS (introduced in Sec.~\ref{sec:mdd-review}) computes the similarity between two time series. From the UCR's SonyAIBORobotSurface1 dataset, time series indexed 82 is queried (green), and is matched with time series indexed 4 (red). Their corresponding envelopes created by $\varepsilon \in \{1, 2, 4, 8, 16\}$ is shown, individually (rows 1 and 2), as well as overlaid (row 3). One example of alignment (where $\varepsilon = 2$) is magnified for better visibility. The union and intersection used in Eq.~\ref{eq:union-and-inters} are marked.}
        \label{fig:mdd_main_plot}
        \vspace*{-0.3cm}
    \end{figure}
    
    The principal objective of this study is to identify and address a specific gap in measuring proximity in time series data. We explore the potential of a recently proposed distance metric, the Multiscale Dubuc Distance (MDD) \cite{khazaei2024multiscale}. We believe this metric can enrich the landscape of distance measures by complementing the most popular ones, namely, Euclidean Distance (EuD), Dynamic Time Warping (DTW), and their derivatives. Accordingly, this study investigates the strengths and weaknesses of MDD in comparison to EuD and DTW.

    To locate the gap that MDD fills, let us begin with a broad generalization about EuD and DTW measures. EuD-inspired measures are typically used either (1) as a first choice when the search task is fairly simple, (2) as a fast choice when computation time is a significant concern, or (3) as a last resort when more advanced measures fail to capture patterns beyond what EuD can detect. DTW-based measures, on the other hand, rely on an elastic definition of similarity, that is, how well the peaks and valleys of two time series can be aligned. When DTW fails to effectively identify the peaks or struggles to match them meaningfully, it often regresses to EuD, unless unsuitable window sizes are used, which can result in pathological warping.

    MDD is defined entirely differently from all other similarity measures, including EuD and DTW (see Sec.~\ref{sec:mdd-review}). The notion of similarity underlying MDD is inspired by object similarity, specifically the Intersection-over-Union (IoU) measure, which is the most popular metric for comparing objects. This inspiration is discussed in \cite{ahmadzadeh2023tsmiou}, where the first version of the MDD measure is introduced as a time series similarity metric without mapping. Comparing time series at different scales (i.e., ``resolutions'', so to speak) without any mapping of values is indeed a novel approach and merits further investigation.

    We fully recognize the multifaceted nature of quantifying similarity/distance between time series. Therefore, we neither intend nor believe it is desirable to pursue a one-size-fits-all distance measure suitable for all applications. The notion of similarity is inherently subjective and should be treated as such. In this study, we target a specific and visible gap, and our findings should be interpreted in that context.

\section{Background}


    As mentioned earlier, similar time series can be defined in different ways, as the notion of similarity differs across different domains. To this end, we will first review how similarity can be defined in different ways.
    
     Euclidean distance is one of the popular generic measures for finding similarity/distance between time series, as the comparison is the most intuitive. EuD does a one-to-one (static) mapping of all the points in a pair of time series and penalizes each mismatch without considering warping.
     
     DTW is a similarity measure that is invariant to warping, which is a desirable feature in many cases and thus, has made it one of the most commonly used measures for classification and clustering of time series \cite{berndt1994using, keogh2002exact, sakoe1987dynamic}. This measure finds the optimal mapping between the two time series in a window $w$ using Euclidean Distance. When $w$ is set to 0, DTW and EuD are equivalent. In the most extreme case, the window size could be equal to the length of the time series. In this case, we say DTW becomes \textit{unconstrained}, meaning that it can map any point on one time series to any point on the other. This highlights the importance of optimizing this parameter, as an unnecessarily large $w$ results not only in pathological warping but also in redundant computations \cite{sakoe1987dynamic,dau_optimizing_2018}. 
     
     In the vast majority of cases, constraining the warping window is essential to achieving high accuracy. 
     The two most established warping paths used to constrain DTW are the Sakoe-Chiba Band \cite{sakoe1987dynamic} and the Itakura Parallelogram \cite{itakura1975minimum}. The Sakoe-Chiba band with a 10\% warping window is the most common choice. This band limits the path to a mostly uniform width around the time series, narrowing at peaks and valleys. The Itakura Parallelogram forms an even tighter band around a timeseries.
     
     Other variants of DTW have been proposed to extend its applications and remedy the possible pathological mapping issues. DDTW uses the first derivative of the time series to find the peaks and valleys; hence a more intuitive approach was introduced that focuses on the shape as well to minimize the misalignments \cite{keogh_derivative_2001}. Another similar measure, shapeDTW, aligns points using DTW based on their local descriptors (e.g., slope, Discrete Wavelet Transform, or Piecewise Aggregate Approximation) in a fixed window to map the points between the two time series. SegDTW is another variant of DTW that avoids pathological warpings by ensuring that DTW first aligns similar features like peaks and valleys by segmenting the time series \cite{ma2018segmentation}. This approach allows for more intuitive alignments and avoids mismatching large sections of the time series while keeping the flexibility of DTW. The versatility and flexibility of DTW make it suitable to be used as a loss function in deep neural networks in its differentiable form, SoftDTW, introduced in \cite{Cuturi2017softdtw}.

     It can be seen how considering the shape of time series can be intuitive to the human eye, and this has inspired other measures that look at the shape rather than mapping points, especially for subsequence matching. Agrawal et al. proposed Shape Definition Language (SDL) where, given a query pattern, the sub-pattern matching that query is retrieved using hierarchical indexing \cite{agrawal_querying_1995}. However, this approach may occasionally misclassify shapes as being similar and is computationally expensive \cite{dong_research_2006}. Another shape-based measure is SpADe \cite{chen_spade_2007}, which uses local pattern matching to remedy shifting and scaling in temporal and amplitude of time series, which existing metrics like Euclidean distance, DTW, LCSS, and EDR struggle with. SpADe takes many parameters as input that need to be fine-tuned, namely, temporal and amplitude scales, the number of local patterns, the largest allowed gap, the penalty function on the gaps between local patterns, and the coefficient of the penalty function. A very similar work to SpADe is AMSS \cite{nakamura2013shape}, which represents time series as vector sequences rather than just data points, allowing it to handle spatial variation better and focus on the sequence's shape using vector directions. The similarity of vectors is then computed using cosine similarity. As AMSS is susceptible to short-term oscillations, the authors suggested using ensemble learning with various smoothing algorithms or similarity measures to address this issue.
     
    Time series as high-dimensional data have microscopic and macroscopic features related to intricate changes and the overall trend of the time series, respectively. To this end, many works have been proposed for capturing information from time series at different granularity levels. Wavelet transform is one tool that allows analyzing the microscopic trends present in the time series.
     One of the most used translated wavelets for finding the similarity of time series is the Haar wavelet used in Discrete Wavelet Transform (DWT) for dimensionality reduction \cite{chan_efficient_1999,struzik_haar_1999}. The Piecewise Aggregate Approximation \cite{keogh_dimensionality_2001} method is also similar to the Haar wavelet in its approach to decomposing and reconstructing the time series and takes advantage of the \textit{local view} that wavelets provide, thanks to being localized in time. 
     
     More recently, with new advances in Deep Neural Networks (DNNs), methods of representing time series in a higher dimension have gained popularity. These techniques can be used for time series generation, but one challenge is ensuring that the output and queried time series remain similar when converted to the original space, for which different techniques have been proposed. T-Loss \cite{franceschi_unsupervised_2019} and TS-TCC \cite{eldele_time-series_2021} are approaches that are inspired by contrastive learning.
        
    Some works consider what features in time series make a similarity measure more suitable for a dataset \cite{wang_rule_2009}; however, in this work, our goal is to discuss how the notion of similarity can be used as a predictor of a measure's failure or success for time series classification. Our motivation is to guide the community on ways to intuitively determine before trying one measure (specifically MDD and DTW), which will likely be more suitable for their task.

\section{Multiscale Dubuc Distance (MDD): A Review}\label{sec:mdd-review}

    Let $\mathcal{X}$ be a set of time series and $\mathcal{Y}$ be a set of labels. We denote a dataset of $n$ labeled time series as $\{(x_i, y_i)\}_{i=1}^{n}$ where $x_i$ is a time series whose label is $y_i$. Naturally, we denote the $t$-th observation of the $i$-th time series as $x_{i,t}$. When needed, we denote a time series of length $d$ as $(x_t)_{t=1}^{d}$. When it does not cause confusion, we omit the subscript and denote a time series by lower-case letters, e.g., $x$, $y$, $z$. For multivariate time series we use the notation $x^{(k)}$ indicating the $k$-th variate. 

    The Multiscale Dubuc Similarity (MDS) metric, introduced in \cite{khazaei2024multiscale}, compares two time series at different scales (or ``resolutions''). Inspired by Dubuc's Variation method \cite{dubuc1989evaluating, taylor2019fractal}, the authors use envelopes of different widths representing time series at different resolutions and quantify the result of all pair-wise comparisons into a single value between 0 and 1.

    The envelope of a time series is defined as follows: Given a time series $(x_t)_{t=1}^{d}$ and an arbitrary value $\varepsilon \in \mathbb{R}_{\geq 0}$ we define an upper bound $u_{x,\varepsilon}(t) = \sup_{t' \in R_\varepsilon (t)}x_{t'}$ and a lower bound $l_{x,\varepsilon}(t) = \inf_{t' \in R_\varepsilon (t)}x_{t'}$, where the neighborhood $R_{\varepsilon}(t)$ is defined in Eq.~\ref{eq:mdd_radius}.
    \begin{equation}\label{eq:mdd_radius}
        R_{\varepsilon}(t) = \{s: |t-s| \leq \varepsilon,\; \; s \in [1,d] \}
    \end{equation}

    \noindent The computed upper and lower bounds define an envelope around the trace of the time series as shown in \ref{fig:mdd_main_plot}.    
    The width of an envelope is controlled by $\varepsilon \in \mathcal{E}$; using larger values of $\varepsilon$ results in obtaining thicker envelopes capturing only the macroscopic trends, whereas smaller values of $\varepsilon$ captures the microscopic trends. For $\varepsilon = 1$, no envelope is generated. This multi-scale view of time series is the unique aspect of the MDS measure.

    Given two time series $x$ and $y$, and an arbitrary window size $\varepsilon \in \mathbb{R}_{\geq 0}$, their \textit{Intersection Ratio} is defined in Eq.~\ref{eq:intersection_ratio}. Note that the intersection and union operations are applied to the computed upper and lower bounds for each value of time series, and it should not be equated with the common area-based understanding of those operation. The union and intersection operations are also defined in Eq.~\ref{eq:union-and-inters}.

    \begin{equation}\label{eq:intersection_ratio}
        r(x,y, \varepsilon) = \frac{\cap_{\varepsilon}(x,y)}{\cup_{\varepsilon}(x,y)}
    \end{equation}
    
    \begin{align}
        \begin{split}
            \cap_{\varepsilon}(x, y) &= \sum_{t = 1}^{d} \max \bigl( \min(u_{x, \varepsilon}(t), u_{y, \varepsilon}(t)) \\
            &\quad\quad\quad\quad\quad - \max(l_{x, \varepsilon}(t), l_{y, \varepsilon}(t)), 0 \bigr), \\
            \cup_{\varepsilon}(x, y) &= \sum_{t = 1}^{d} \max \bigl( \max(u_{x, \varepsilon}(t), u_{y, \varepsilon}(t)) \\
            &\quad\quad\quad\quad\quad - \min(l_{x, \varepsilon}(t), l_{y, \varepsilon}(t)), 0 \bigr).
        \end{split}\label{eq:union-and-inters}
    \end{align}

    The final step for calculating MDS is to aggregate all values of $r(x,y, \varepsilon)$, for $\varepsilon \in \mathcal{E}$. To do so, MDS, as formulated in Eq.~\ref{eq:mds}, calculates a pseudo-area under the curve of $r(x,y, \varepsilon)$ for $\varepsilon \in \mathcal{E}$.
        
    \begin{equation}
        \text{MDS}(x,y, \mathcal{E}) = \sum_{i=2}^{\mathcal{|E|}} \frac{r(x, y, \varepsilon_{i-1}) + r(x, y, \varepsilon_i)}{2} \Delta\varepsilon_i
        \label{eq:mds}
    \end{equation}

    Since $r(x,y, \varepsilon) \in [0,1]$, with a proper choice of $\Delta\varepsilon$, the area under the curve of the intersection ratio is also confined between 0 and 1. This choice is $\Delta\varepsilon_i = \frac{1}{|\mathcal{E}|}$. In cases where two time series match perfectly, i.e., $\forall t, x_t = y_t$, we achieve $\text{MDS}(x,y, \mathcal{E}) = 1$. Note that Eq.~\ref{eq:mds} computes pseudo-area, meaning that the actual values of $\varepsilon$ (i.e., powers of 2) do not determine $\Delta\varepsilon_i$.

    \textbf{Key Features.} Because $\forall x, y, \varepsilon, \text{MDS}(x,y, \mathcal{E}) \in [0,1]$, it can serve as a distance measure as well. The \textit{Multiscale Distance Measure} (MDD) is defined as $\text{MDD}(x,y, \mathcal{E}) = 1 - \text{MDS}(x,y, \mathcal{E})$. It is shown that MDD is indeed a (pseudo-) \textit{metric}. That is, it holds the following properties for any time series $x$, $y$, and $z$:

    \begin{itemize}
        \item $\forall x, \; \text{MDD}(x,x) = 0 \;\;$ (reflexivity)
        \item $\forall x, y, \; \text{MDD}(x,y) \geq 0 \;\;$ (positiveness)
        \item $\forall x, y, \rm{if}\;x \neq y, \; \text{MDD}(x,y) > 0 \;\;$ (strict positiveness)
        \item $\rm{if}\;x \neq y, \; \text{MDD}(x,y) = \text{MDD}(y,x) \;\;$ (symmetry)
        \item $\forall x,y,z, \; \text{MDD}(x,z) \leq \text{MDD}(x,y) + \text{MDD}(y,z)\;\;$ (triangle inequality)
    \end{itemize}

    Further, as shown in \cite{khazaei2024multiscale}, the running time of the algorithm for computing MDD is linear in the length of time series. To be exact, its running time belongs to the class $\Theta(|\mathcal{E}| \cdot d)$ where $|\mathcal{E}|$ is a constant (i.e., the number of different choices of $\varepsilon$) and $d$ is the length of the time series.

\section{Datasets and Evaluation Framework}\label{sec:data}

    In this study, we investigate the strengths and weaknesses of MDD, first on the UCR Time Series Classification Archive \cite{UCRArchive2018}, and then on a real-world dataset, namely Space Weather Analytics for Solar Flares (SWAN-SF) \cite{angryk2020multivariate}. The UCR archive of 128 univariate time series datasets, is shown to be a valuable resource for the data mining and machine learning community with more than one thousand citations in Google Scholar. Further, several publications by the team who collected and curated the dataset outlines frameworks to distinguish (as much as possible) good practices and reliable findings from anecdotal and non-reproducible ones.
    
    For our experiments, we excluded 11 datasets containing time series of unequal lengths, as varying-length sequences are treated differently by various similarity measures, which could introduce confounding effects. Additionally, 22 datasets with very long time series (exceeding 900 observations) were omitted, based on the rationale that analyzing full-length sequences of such size has limited practical relevance. The performance of a similarity measure on these extended time series offers little meaningful insight into its capabilities.

    The SWAN-SF dataset is a machine learning benchmark dataset created for space-weather forecasting community to unify the flare forecasting efforts and make the findings comparable. Solar flares are emissions of electromagnetic radiation in the Sun's atmosphere. Extreme solar flares are known to cause catastrophic damage to our infrastructure. This dataset contains over four thousand multivariate time series, monitoring 51 flare-predictive parameters, between 2010 and 2018. Each time series has 60 records of magnetic-field parameters, collected within a window of 12 hours. Furthermore, each observation in this dataset overlaps by up to 11 hours. As mentioned in \cite{angryk2020multivariate} the authors of this dataset have accounted for this temporal coherence by pre-partitioning the data into five non-overlapping time periods, each containing roughly the same number of extreme flaring instances. The imbalance ratio varies between partitions.

    For our experiments, we preprocess the dataset as recommended by domain experts in \cite{ahmadzadeh2021how}. Primarily, preprocessing must account for both class imbalance and temporal coherence. To preserve temporal coherence we follow the recommendation of the authors and use separate partitions, for train and test pairs. To address the class imbalance, we follow the \textit{climatology-preserving sampling} strategy, recommended in \cite{ahmadzadeh2021how}: the majority class is undersampled to achieve balance, while the original climatology of flares within each class is maintained. More precisely, SWAN-SF divides flare magnitudes into five classes of intensity $X$, $M$, $C$, $B$, and $N$, from strongest to weakest. We group these classes to form two super classes, namely Flaring ($FL = \{X,M\}$) and Non-flaring ($NF = \{C,B,N\}$). We use random undersampling to achieve a balance of $FL$ and $NF$ instances while preserving the subclass ratios. 
    Finally, since the focus of this study is not on flare forecasting, we simplified this data set by reducing its feature space down to only three features. We select the top ranked features with most predictive power as reported in several studies including \cite{yolekar2021feature, bobra2015solar}. The utilized features are named ``TOTUSJH", ``TOTBSQ", and ``TOTPOT". For their definition, please see the original study.

\section{Experiments and Results}

    \subsection{MDD vs DTW (with Optimized Parameters)}\label{sec:mdd_dtw}
        To provide an overview as to how MDD compares with DTW, we repeat the experiment carried out in \cite{khazaei2024multiscale} with one major difference: previously, while DTW was used with an optimal (learned) window size per dataset, MDD was not---a fixed set $\mathcal{E}$ was used for all datasets. In this experiment, we optimize the window size of MDD as well.
        
        Our experimental setting is as follows: We use the 95 datasets of the UCR archive (discussed in Sec.~\ref{sec:data}) and run the 1-Nearest Neighbors (1NN) algorithm to classify each time series using MDD as the distance measure. Then, we compare its classification performance with that of 1NN with DTW as the distance measure. This allows a fair comparison of the effectiveness of the two metrics, DTW and MDD. To optimize MDD, for each dataset we search for a set $\mathcal{E}$ which maximizes the classification accuracy of the training set. We define our search space as all powers of two, which are less than or equal to a pre-defined fraction of the length of time series in that dataset. In other words, our search space is the power set $2^S$ where $S = \{2^i; 2^i \leq \eta d\}$, $d$ is the length of time series in a dataset, and $\eta$ is the chosen fraction. For example, for $\eta\!=\!0.3$ and time series of length $d = 200$, we determine $S$ to be $\{1, 2, 4, \dots, 32\}$, because $2^5$ is the largest power of two less than or equal to $(0.3)(200) = 60$. The choice of $\eta$ prevents the largest envelopes to be unreasonably thick. Note that, $\varepsilon$ is the radius (not diameter; see Eq.~\ref{eq:mdd_radius}), so in the numerical example given, $\eta=0.3$ allows envelopes to be constructed by rolling windows as large as $2 \times 32$ which is already very large for time series of length 200---in most cases, this window size is not optimal.

        \begin{figure}
            \centering
            \includegraphics[width=0.8\linewidth]{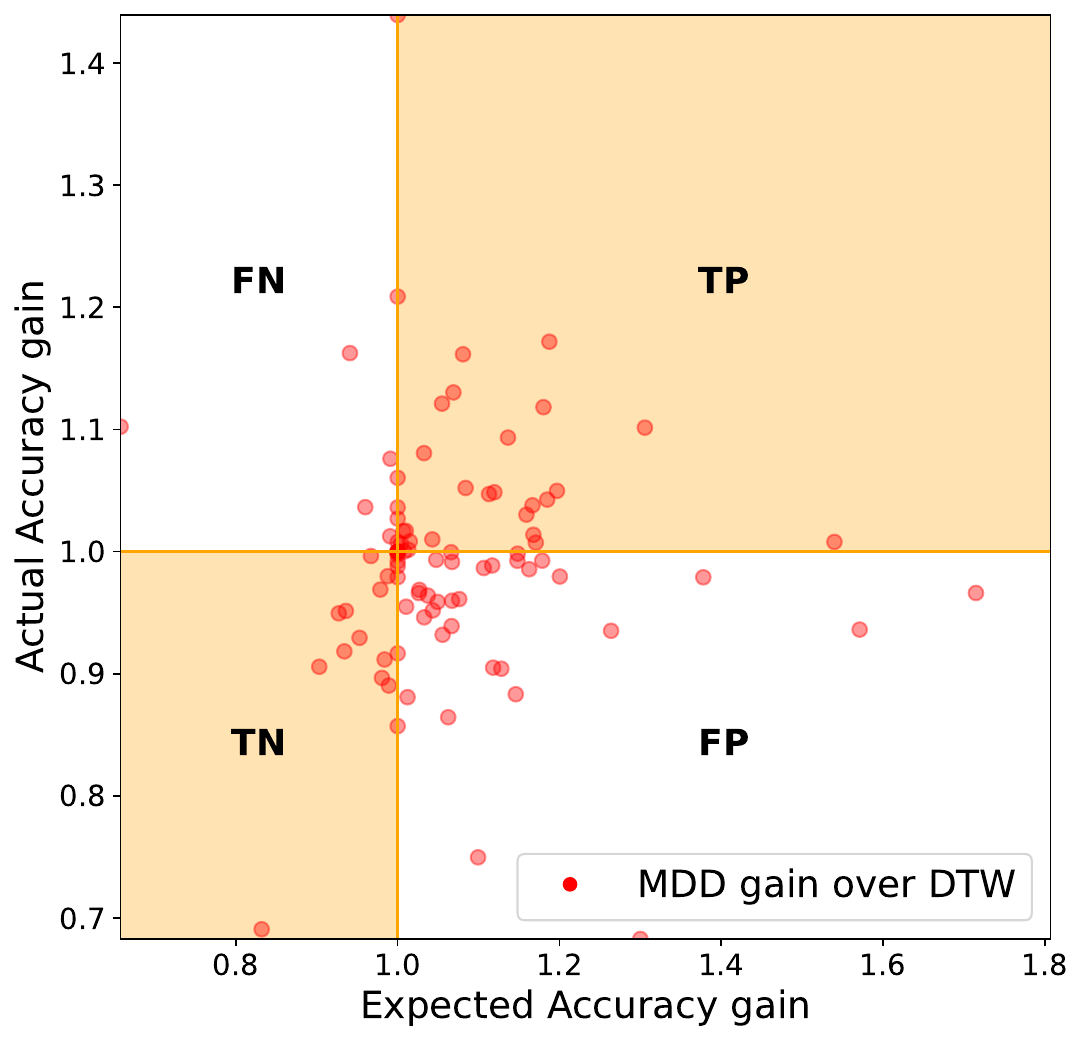}
            \caption{The Sharpshooter plot compares the actual and expected accuracy gain of MDD over DTW ($g_{mdd,dtw}$, $\hat{g}_{mdd,dtw}$, respectively) for 95 datasets of UCR.}
            \label{fig:sharpshooter1}
        \end{figure}

        The results of this experiment is shown in Fig.~\ref{fig:sharpshooter1} via the Texas Sharpshooter plot, that is the evaluation framework recommended by the creators of the UCR dataset \cite{batista2011complexity, dau2019ucr, UCRArchive2018}. The Texas Sharpshooter plot is a 2-dimensional plot in which each point represents the performance gain of one distance measure relative to another. More specifically, for each dataset, we run the 1NN classifier on the training set using leave-one-out cross-validation, and classify each time series. Denoting the utilized distance measure by $\mu$, the accuracy of this run is called the \textit{expected accuracy}, $\widehat{acc}_\mu$. We then run 1NN again, but this time to classify the time series of the test set, by measuring their proximity to the time series in the training set. This is called the \textit{actual accuracy}, ${acc}_\mu$. To compare the relative contribution of different distance measures $\mu_1$ and $\mu_2$, the \textit{expected accuracy gain} ($\hat{g}_{\mu_1, \mu_2}$) and \textit{actual accuracy gain} ($g_{\mu_1, \mu_2}$) are defined in Eq.~\ref{eq:gain}. Fig.~\ref{fig:sharpshooter1} shows $\hat{g}_{\text{mdd}, \text{dtw}}$ and $g_{\text{mdd}, \text{dtw}}$, for 95 datasets. Note that the training and test sets for each dataset are already separated in the UCR archive.

        \begin{equation}\label{eq:gain}
            \hat{g}_{\mu_1, \mu_2} = \frac{\widehat{acc}_{\mu_1}}{\widehat{acc}_{\mu_2}}, \;\;
           g_{\mu_1, \mu_2} = \frac{{acc}_{\mu_1}}{{acc}_{\mu_2}}
        \end{equation}

        MDD outperforms DTW in 41\% of the UCR datasets. This is 14\% higher than what was initially reported in \cite{khazaei2024multiscale} where $\mathcal{E}$ was not optimized. 34\% of the datasets are in the TP region meaning that for them MDD was expected to outperform DTW and it did. The plot also shows MDD's gain is overestimated (in FP region) for 36\% of datasets and underestimated (in FN region) for 5\% of datasets. As explained in the introduction, the intention of this analysis is not to assert any measure as best in every use case, but instead to identify and demonstrate the specific cases to which MDD or DTW is best suited. We investigate exactly that in the following subsections.

\begin{figure}
            \centering
            \includegraphics[width=1\linewidth]{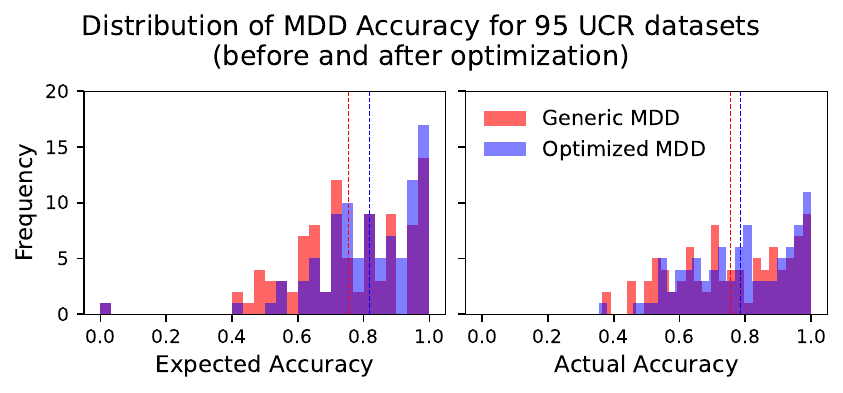}       
            \caption{Comparison of optimization approaches. These figures show the distribution of accuracy scores before and after optimizing MDD. The expected accuracy improves from 0.75 to 0.82, and the actual accuracy improves from 0.75 to 0.78. In the expected phase we show only the optimal accuracy scores.}
            \label{fig:opt_distro_shift}
        \end{figure}

        Focusing on the improvement of 1NN using optimized MDD, the distribution of the expected and actual accuracy scores, before and after optimization is illustrated in Fig.~\ref{fig:opt_distro_shift}. The average of expected accuracy improves from 0.75 to 0.82, and the average actual accuracy improves from 0.75 to 0.78.

    \subsection{When DTW Is Better}
        Although it might be challenging to come up with an exclusive list of characteristics which render a dataset more suitable for DTW, we can identify the most trivial ones. When the subjective notion of similarity in the problem in hand is defined in terms of the matching of prominent peaks and valleys (and if such features exist in the data), DTW is expected to outperform other measures, or at least, be as good as the best ones, including MDD. This should not come as a surprise because DTW determines similarity by finding an optimal matching of points \cite{using1994berndt}.

\begin{figure}
            \begin{subfigure}{1.0\linewidth}
                \centering
                \includegraphics[width=1.0\linewidth]{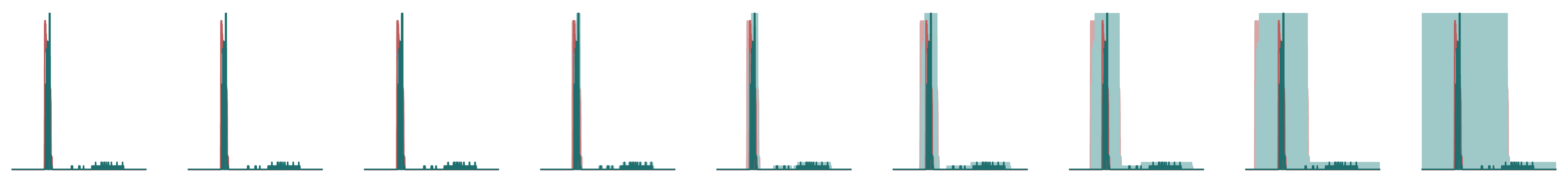}
                \caption{MDD's envelopes for two time series matched incorrectly with $i=11$, by MDD.}
                \label{fig:LK_11_mdd_match}
            \end{subfigure}

            \bigskip
            
            \begin{subfigure}{1.0\linewidth}
                \centering
                \includegraphics[width=1.0\linewidth]{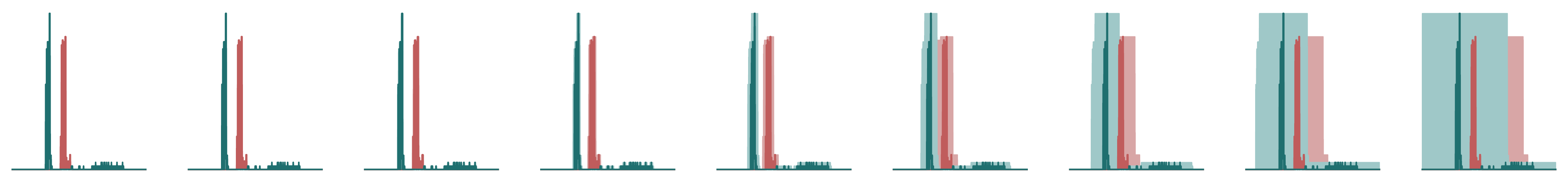}
                \caption{MDD's envelopes for two time series matched correctly with $i=11$, by DTW.}
                \label{fig:LK_11_dtw_match}
            \end{subfigure}

            \bigskip

            \begin{subfigure}{1.0\linewidth}
                \centering
                \includegraphics[width=1.0\linewidth]{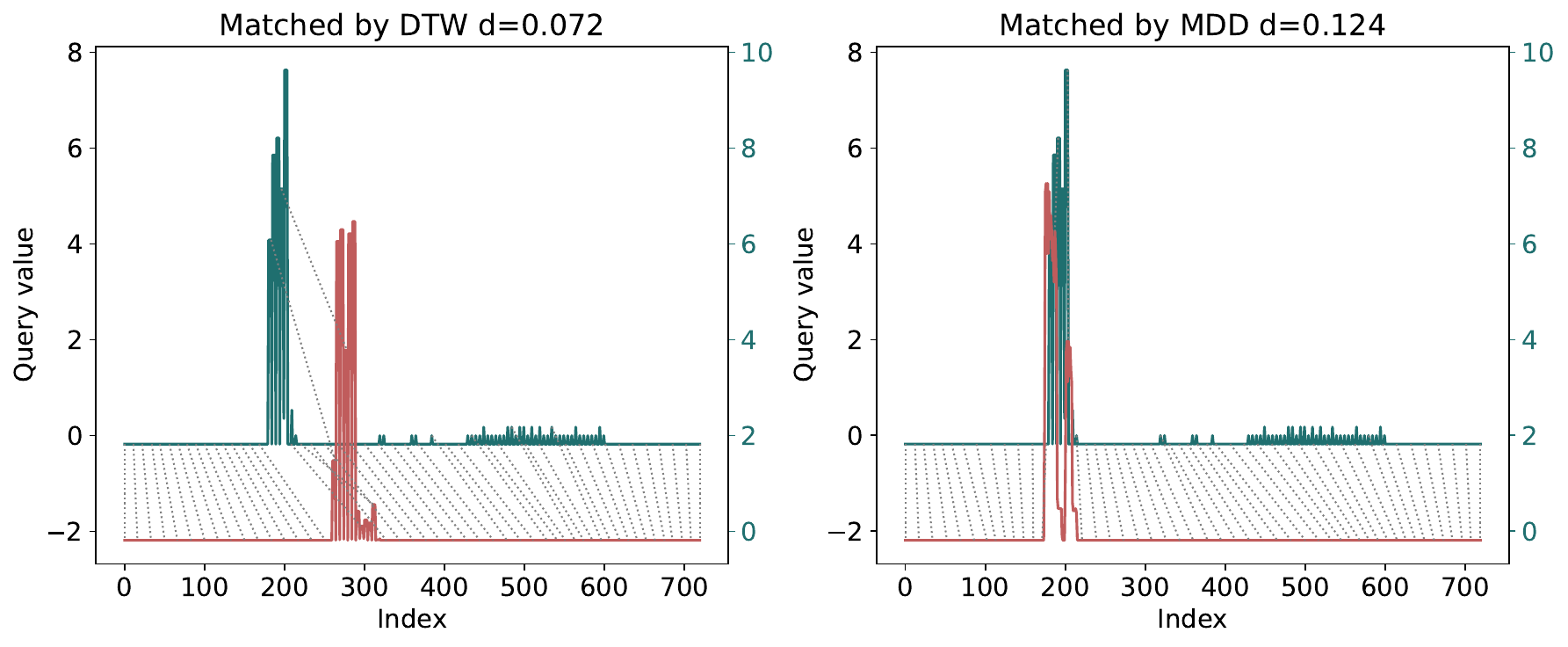}
                \caption{DTW's point matching for time series matched by DTW (left) and MDD (right). Note the effectiveness of DTW and failure of MDD in matching time series in the left plot. To generate this and some other plots, we used the DTW Suite package \cite{JSSv031i07}.}
                \label{fig:LargeKitchenAppliances_11}
            \end{subfigure}
            
            \caption{An example of a pair of time series (from the LargeKitchenAppliances dataset) for which DTW is most suited and MDD fails to capture the similarities.}
            \label{fig:LK_11_bounds}
        \end{figure}

        To illustrate this point, let us take a close look at the LargeKitchenAppliances dataset, as an example. 1NN with DTW achieved an accuracy of $79.4\%$ and outperformed 1NN with MDD with an accuracy of $54.9\%$. This dataset contains three classes of time series of length 720, each of which exhibits outstanding peaks, with distinguishable differences in shapes and numbers. In Fig.~\ref{fig:LK_11_bounds}, two time series of this dataset, belonging to the same class, are illustrated. It is evident in Figs.~\ref{fig:LK_11_mdd_match} and \ref{fig:LK_11_dtw_match}, where MDD's envelopes are illustrated, why MDD fails to capture the similarities; the shift between the outstanding peaks of the two time series is too large and far apart to be captured by MDD's envelopes, whereas DTW, as shown in Fig.~\ref{fig:LargeKitchenAppliances_11} can effectively capture the shift.
        
        The strength of DTW lies in its elasticity directly correlated with its window size; larger windows allow more elastic matching of points. However, this example exhibits only one form of (or definition of) elasticity. MDD is also an elastic measure, as we will see in other cases, but without warping.

        Also, for DTW to be effective for this dataset, it had to use an unusually large window size. Looking at the distribution of all DTW's optimal window sizes, a window size of 94 (used for LargeKitchenAppliances) ranks at the 98th percentile (when normalized by time series lengths). This is 3.8 standard deviations larger than the mean optimized window size for all datasets in the UCR archive. This places LargeKitchenAppliances in the 3rd place, after MedicalImages and ElectricDevices, in terms of the size of the optimized windows. However, the length of time series in those datasets (99 and 96, respectively) is significantly smaller than that of LargeKitchenAppliances (720). This analysis tells us that, DTW has the power to capture such similarities using a large window, and of course, larger windows require more computations for DTW to find an optimal warping path.

        The second group of datasets on which DTW is expected to perform very well exhibits equally large within-class variability.

        \begin{figure}
            \begin{subfigure}{1.0\linewidth}
                \centering
                \includegraphics[width=1.0\linewidth]{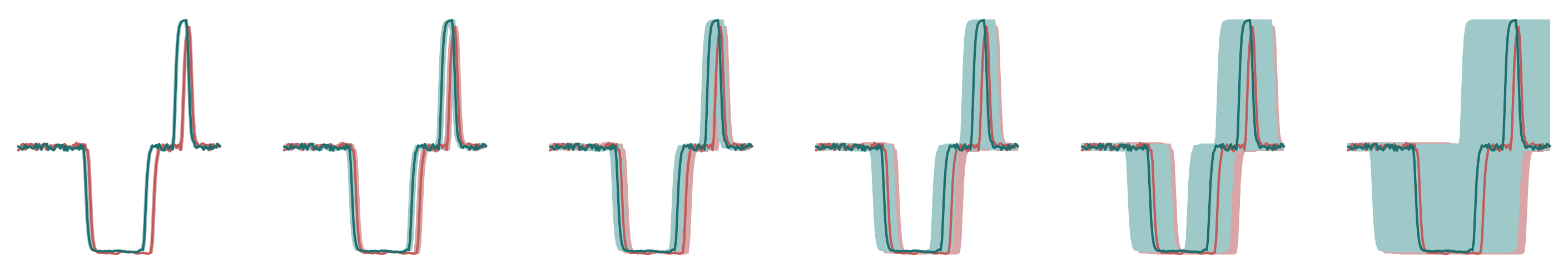}
                \caption{MDD's envelopes for time series matched with $i=16$, by MDD}
                \label{fig:umd_16_mdd_match}
            \end{subfigure}

            \bigskip
            
            \begin{subfigure}{1.0\linewidth}
                \centering
                \includegraphics[width=1.0\linewidth]{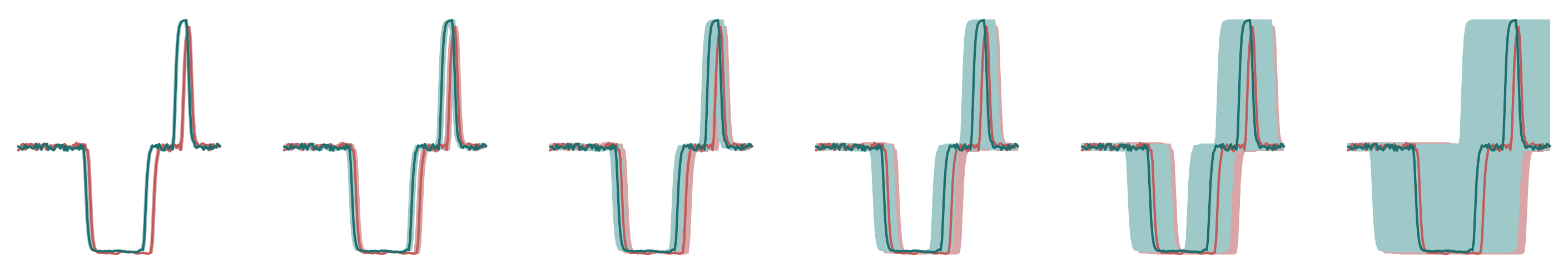}
                \caption{MDD's envelopes for time series matched with $i=16$, by DTW}
                \label{fig:umd_16_dtw_match}
            \end{subfigure}

            \bigskip

            \begin{subfigure}{1.0\linewidth}
                \centering
                \includegraphics[width=1.0\linewidth]{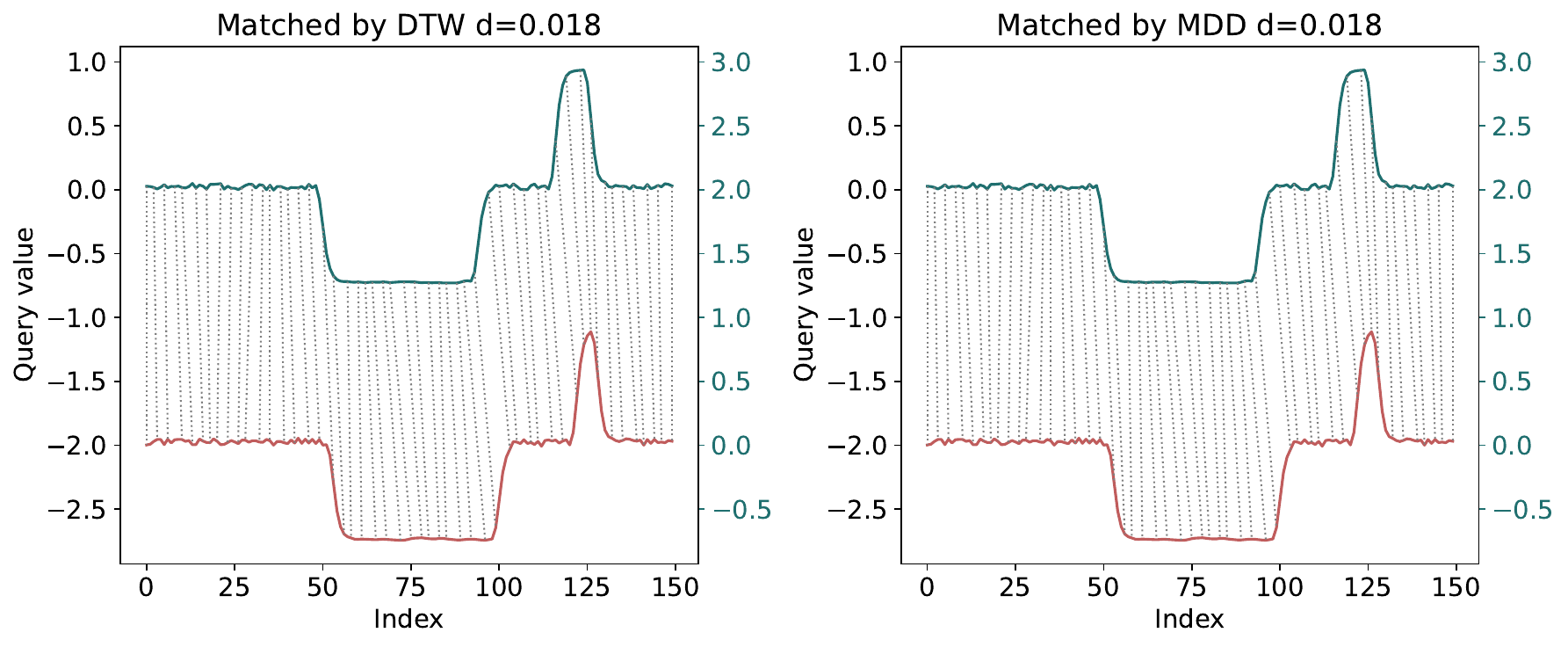}
                \caption{DTW's point matching for time series matched by DTW and MDD, respectively}
                \label{fig:UMD_16}
            \end{subfigure}
            
            \caption{An example of a pair of time series (from UMD dataset) where both metrics perform similarly.}
            \label{fig:UMD_16_bounds}
        \end{figure}
        
    \subsection{When DTW and MDD Tie}
        In several cases the gain is marginal; almost a tie between DTW and MDD. In these cases, an overarching reason is that the within-class variability is much less than the between-class variability. The UMD dataset is a good example of a tie. Looking at a few time series per class one can visually verify the significant difference in the class variability. Looking at the MDD's envelopes and DTW's matching shown in Fig. \ref{fig:UMD_16_bounds}, we can see that the classification task is rather simple, evident by the very high performance of 1NN with DTW or MDD. DTW and MDD achieve almost perfect accuracy of 97\% and 96.5\%, respectively. However, 1NN with EuD achieves a suboptimal accuracy of 76\%, which shows that the variability that we observed is beyond what EuD can capture due to its rigid nature.
                 
        DTW and MDD also tie when they both are rendered ineffective. In such cases, the failure of DTW can be traced back to the fact that its functionality is reduced to EuD. That is, the optimal window size was found to be $w = 0$. A similar reduction happens to MDD, although it is incorrect to say that it is \textit{reduced} to EuD as MDD does not practice a mapping strategy. That said, in similar cases, the optimal set $\mathcal{E}$ for MDD is found to be $\{1\}$. Of course, not every measure has the flexibility to behave like EuD when needed, which can be seen as a strength for MDD. But it is important to note that, DTW reduces to EuD in about 23.4\% of datasets (30 of 128) but MDD becomes ineffective in about 4.21\% of datasets (4 of 95). Below, we can dive deeper in a few cases for further evidence. This discussion raises an important question: how significant are the wins, comparing DTW and MDD? We address this in Sec.~\ref{subsec:significance_of_wins}.
        
        \begin{figure}
            \begin{subfigure}{1.0\linewidth}
                \centering
                \includegraphics[width=1.0\linewidth]{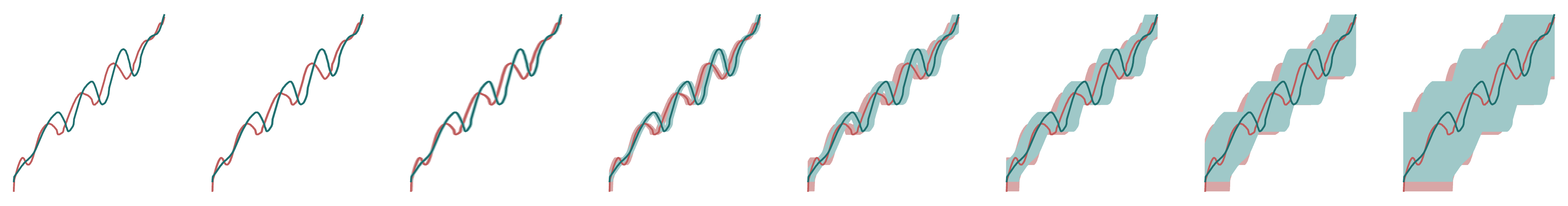}
                \caption{MDD's envelopes for time series matched with $i=114$, by MDD}
                \label{fig:Symbols_114_mdd_match}
            \end{subfigure}

            \bigskip
            
            \begin{subfigure}{1.0\linewidth}
                \centering
                \includegraphics[width=1.0\linewidth]{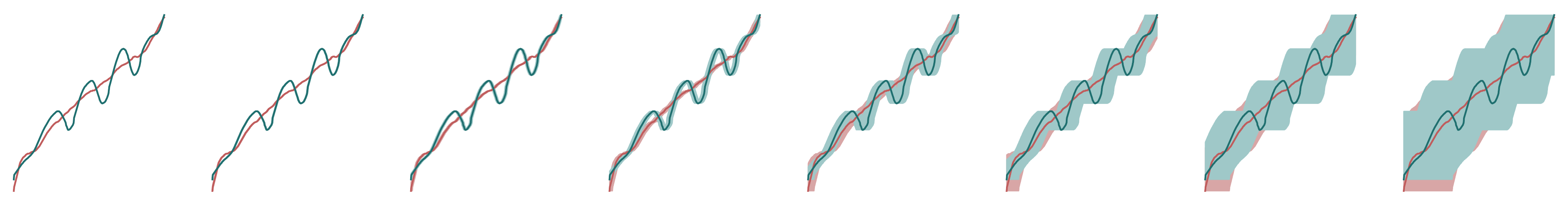}
                \caption{MDD's envelopes for time series matched with $i=114$, by DTW}
                \label{fig:Symbols_114_dtw_match}
            \end{subfigure}

            \bigskip

            \begin{subfigure}{1.0\linewidth}
                \centering
                \includegraphics[width=1.0\linewidth]{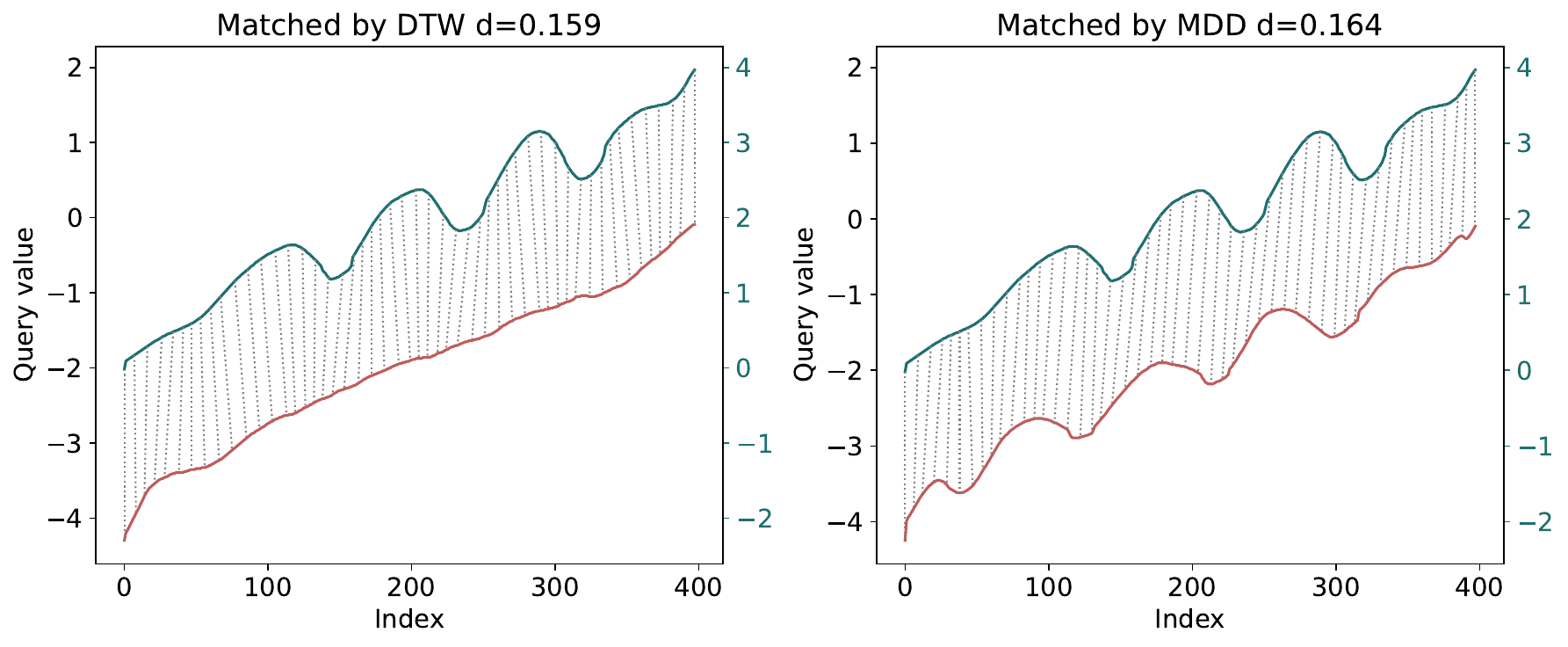}
                \caption{DTW's point matching for time series matched by DTW and MDD respectively}
                \label{fig:Symbols_114}
            \end{subfigure}
            
            \caption{An example of a pair of time series (from the Symbols dataset) where MDD is more suited and DTW fails to capture similarities.}
            \label{fig:Symbols_114_bounds}
        \end{figure}
        
        \begin{figure}
            \centering
            \includegraphics[width=1\linewidth]{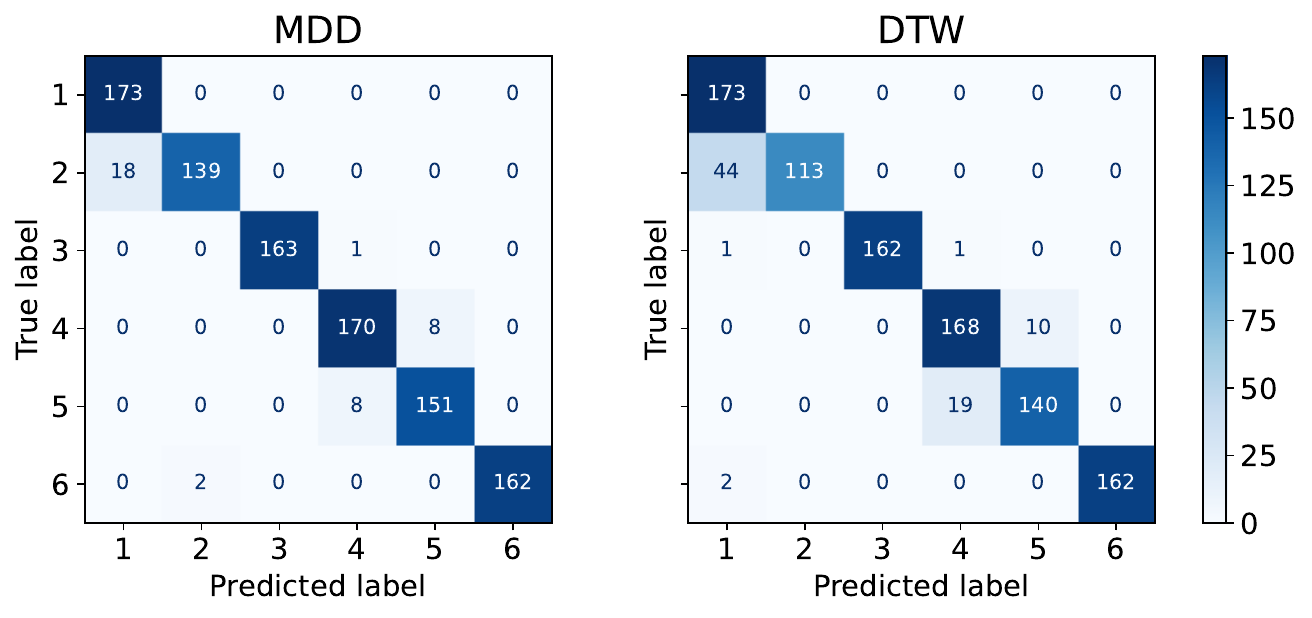}
            \caption{Confusion matrix for 1NN with MDD (left) and DTW (right) on the Symbols dataset.}
            \label{fig:combined_confusion_Symbols}
        \end{figure}

        A dataset showcasing the above argument is the Symbols dataset. MDD and DTW almost tie, and despite their entirely different inner workings, their error originates largely from the same classes. As depicted in Fig. \ref{fig:combined_confusion_Symbols}, DTW shows real struggle in the classification of class-2 instances (misclassified exclusively as class-1; see Fig. \ref{fig:Symbols_114_bounds} as an example), and less so of class-5 instances (misclassified exclusively as class-4). Similarly, MDD shows struggle in classification of class-1 instances (misclassified exclusively as class 2) and of class-4 instances (misclassified exclusively as class-5), and less so of class-6 instances. Neither of the two measures can do much more than EuD about those classes. In Table \ref{tab:Symbols_metrics} we report all performances for classification of each of those classes versus the rest.

        \begin{table}[]
        \centering
            \begin{tabular}{|l|l|l|l|l|l|l|l|}
            \hline
            \textbf{Method}               & \textbf{Class} & \textbf{TP}  & \textbf{FP} & \textbf{FN} & \textbf{Precision} & \textbf{Recall} & \textbf{F1 Score} \\ \hline
            \multirow{4}{*}{\textbf{DTW}} & 1     & 173 & 47 & 0  & 0.79      & 1.0    & 0.88     \\ \cline{2-8} 
                                 & 2     & 113 & 0  & 44 & 1.0       & 0.72   & 0.84     \\ \cline{2-8} 
                                 & 4     & 168 & 19 & 10 & 0.89      & 0.94   & 0.92     \\ \cline{2-8} 
                                 & 5     & 140 & 10 & 19 & 0.93      & 0.9    & 0.91     \\ \hline
            \multirow{4}{*}{\textbf{MDD}} & 1     & 173 & 18 & 0  & 0.91      & 1.0    & 0.95     \\ \cline{2-8} 
                                 & 2     & 139 & 2  & 18 & 0.99      & 0.89   & 0.93     \\ \cline{2-8} 
                                 & 4     & 170 & 9  & 8  & 0.95      & 0.95   & 0.95     \\ \cline{2-8} 
                                 & 5     & 151 & 8  & 8  & 0.95      & 0.95   & 0.95     \\ \hline
            \end{tabular}
        \caption{A comparison of DTW and MDD on classes 1, 2, 4, and, 5 (versus the rest) of the Symbols dataset. True Positive (TP), False Positive (FP), and False Negative (FN) numbers are presented along with  $\text{Precision} = \frac{TP}{TP+FP}$ , $\text{Recall}= \frac{TP}{TP+FN}$, and F1 Score which is the harmonic mean of Precision and Recall.}
        \label{tab:Symbols_metrics}
        \end{table} 

    \subsection{Comparing significance of wins}\label{subsec:significance_of_wins}
        In Sec.~\ref{sec:mdd_dtw} we presented statistics of wins for DTW and MDD. It would be more informative if we talked about---not just the number of wins for each measure, but---the significance of the wins as well. We can do this simply by computing the sum of the (Euclidean) distances of each point on the Texas Sharpshooter plot from the no-gain point, i.e., $\mathbf{1} = (1.0, 1.0)$. Specifically, we compute the significance of wins using Eq.~\ref{eq:sig_win}, once for all datasets where DTW outperforms MDD (i.e., $\mu_1 = \text{DTW}$, $\mu_2 = \text{MDD}$), and then for all datasets where MDD outperforms DTW (i.e., $\mu_1 = \text{MDD}$, $\mu_2 = \text{DTW}$).

        \begin{align}\label{eq:sig_win}
        {\rm{win}}_{\mu_1, \mu_2} &=   \| (g_{\mu_1, \mu_2}, \hat{g}_{\mu_1, \mu_2}) - \mathbf{1} \|_2 , \\
          & \;\; \rm{for}\;\; g_{\mu_1, \mu_2}, \hat{g}_{\mu_1, \mu_2} \geq 1.0 \nonumber
        \end{align}

\begin{figure}
                \centering
                \includegraphics[width=0.9\linewidth]{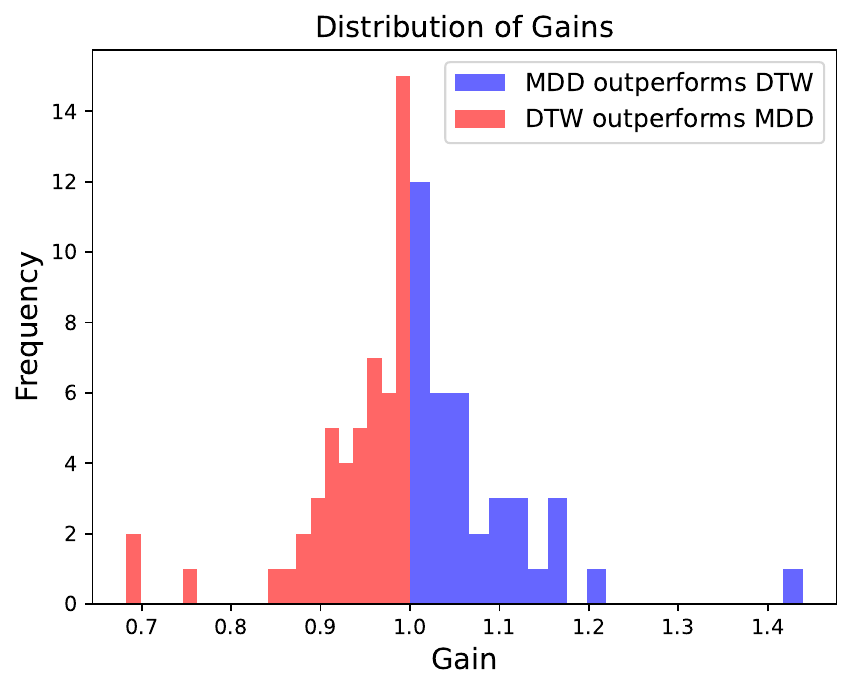}
                \caption{The distribution of actual performance gain on 93 datasets (excluding ties), divided into two parts: when DTW wins (red) and when MDD wins (blue).}
                \label{fig:gains_dist}
        \end{figure}

        When MDD outperforms DTW $\text{win}_{\text{MDD}, \text{DTW}}= 5.86$ ($\mu = 0.15, \sigma = 0.17$), whereas, for cases in which DTW wins $\text{win}_{\text{DTW}, \text{MDD}}= 6.89$ ($\mu = 0.13, \sigma = 0.14$). The mean values normalize for the difference between the number of wins per group. This small gap between the means tells us that the wins by DTW and MDD are equally significant. In other words, no measure is winning more marginally. Fig.~\ref{fig:gains_dist}, illustrates the distribution of gains.
  
        \vspace*{-0.2cm}

    \subsection{When MDD Is Better}\label{sec:sine_cosine_exp}

        In 41\% of datasets, MDD outperforms DTW. We believe that there are several scenarios making a dataset more suitable for MDD. Since MDD avoids point matching to measure the similarity, it is theoretically ideal for datasets where the matching strategy fails; the notion of similarity is not defined through matching of points.

\begin{figure}
            \begin{subfigure}{.49\linewidth}
                \centering
                \includegraphics[width=1.0\linewidth]{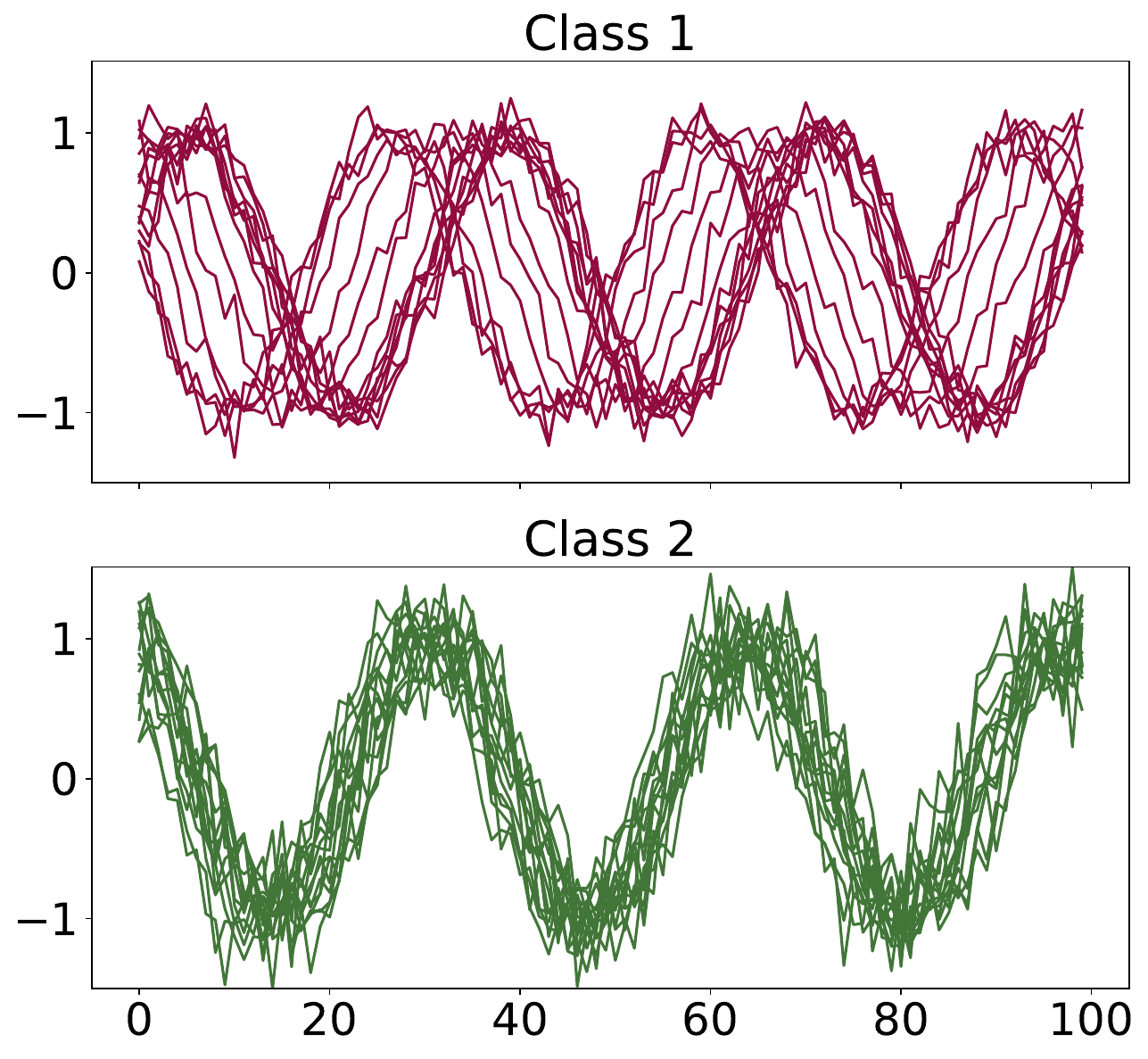}
                \caption{Dataset: Simulated; \\ $acc_{\text{DTW}} = 0.59$, $acc_{\text{MDD}} = 0.77$}
                \label{fig:sine_cosine_dataset2_sim}
            \end{subfigure}
            \begin{subfigure}{.5\linewidth}
                \centering
                \includegraphics[width=1.0\linewidth]{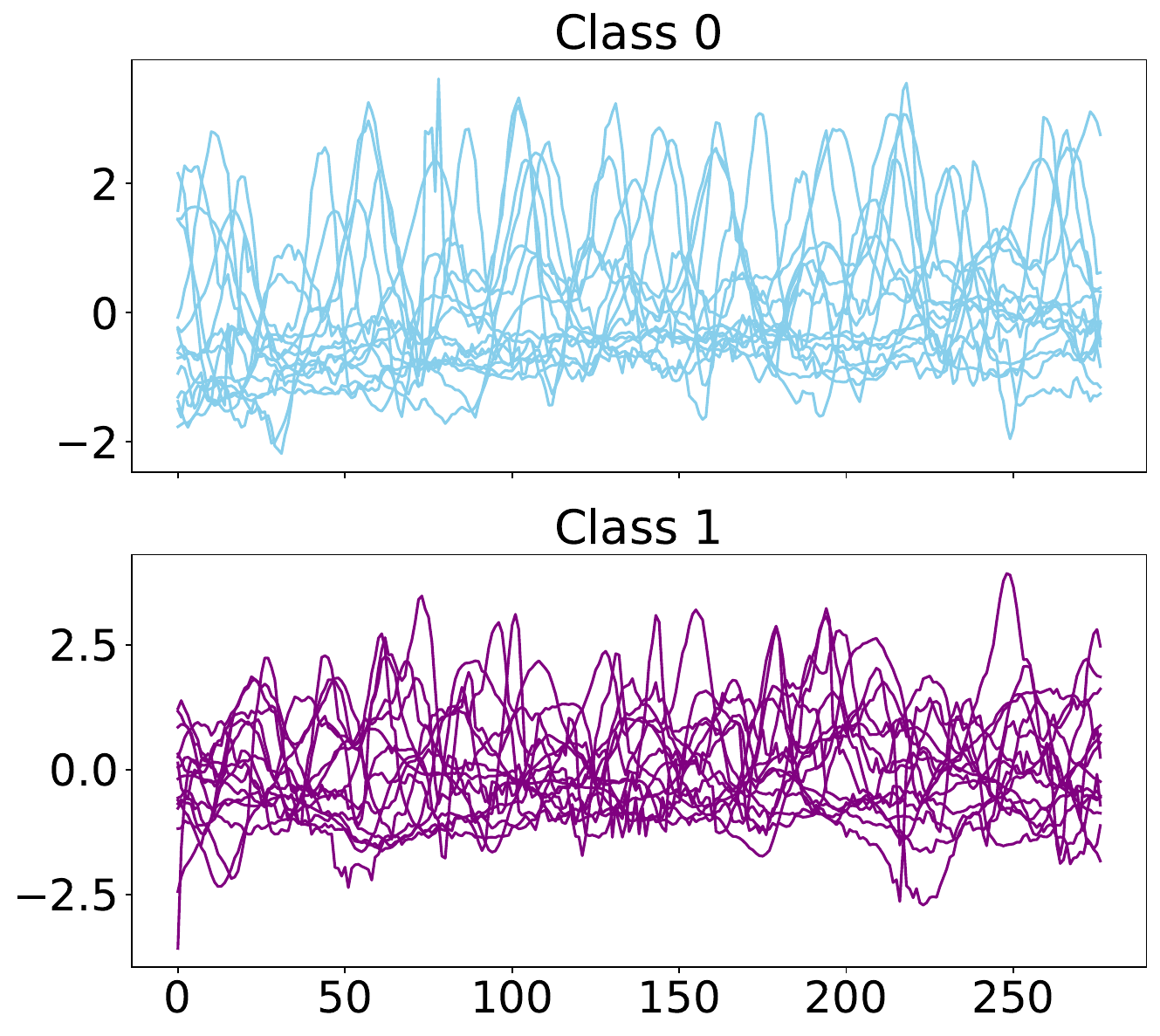}
                \caption{Dataset: ToeSegmentation1; $acc_{\text{DTW}} = 0.75$, $acc_{\text{MDD}} = 0.82$}
                \label{fig:ToeSegmentation1_viz}
            \end{subfigure}

            \bigskip
                        
            \begin{subfigure}{.49\linewidth}
                \centering
                \includegraphics[width=1.0\linewidth]{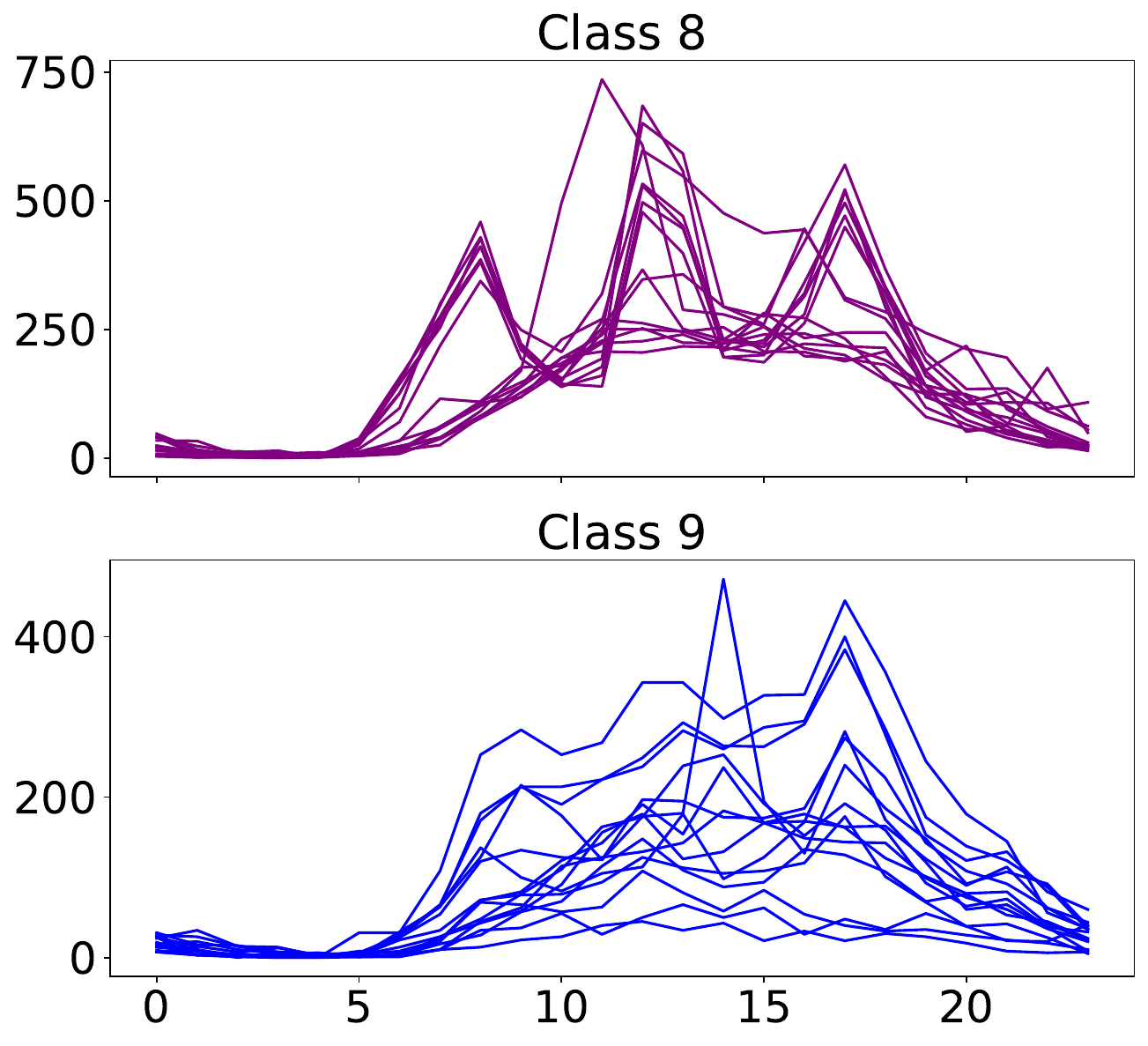}
                \caption{Dataset: MelbournePedestrian; \\ $acc_{\text{DTW}} = 0.81$, $acc_{\text{MDD}} = 0.94$}
                \label{fig:MelbournePedestrian_viz}
            \end{subfigure}
            \begin{subfigure}{.5\linewidth}
                \centering
                \includegraphics[width=1.0\linewidth]{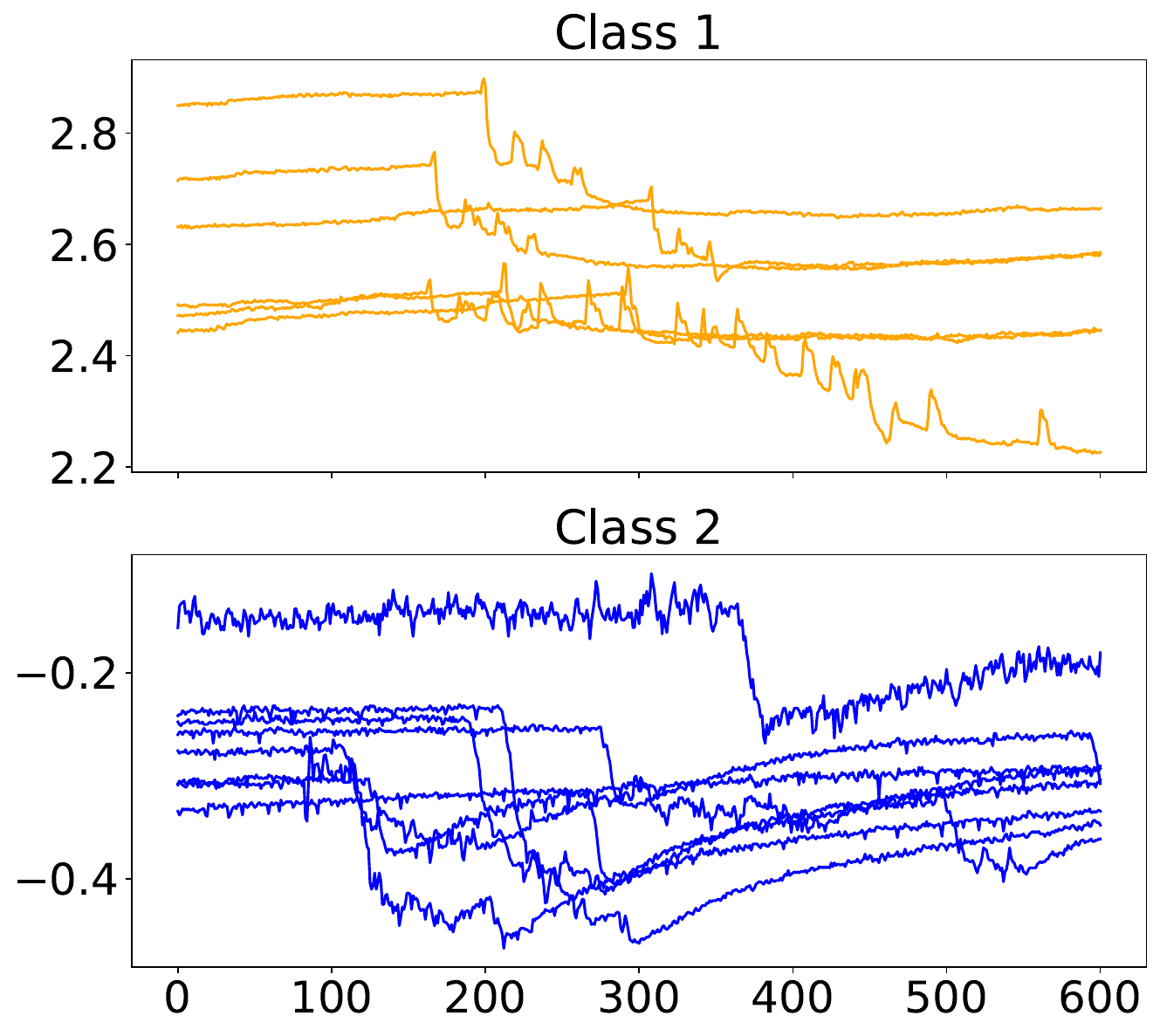}
                \caption{Dataset: InsectEPGSmallTrain; \\ $acc_{\text{DTW}} = 0.69$, $acc_{\text{MDD}} = 1.00$}
                \label{fig:InsectEPGSmallTrain_viz}
            \end{subfigure}
            
            \caption{Examples of UCR dataset in which even though a clear pattern of peaks and valleys is present, DTW falls short when similarity is not defined via point matching.}
            \label{fig:phase_shift}
        \end{figure}

        Hypothetically, let's suppose a dataset contains two classes, and while time series in each class exhibit clear peaks, the phase variance in one class makes it difficult for DTW to effectively discriminate between them by matching points. A few datasets of UCR exhibit this very pattern, such as ToeSegmentation, MelbournePedestrian, and InsectEPGSmallTrain. Examples of those time series are illustrated in Figs.~\ref{fig:ToeSegmentation1_viz}, \ref{fig:MelbournePedestrian_viz}, \ref{fig:InsectEPGSmallTrain_viz}. In all those cases MDD outperforms DTW. 
        
        To generalize our hypothesis, we simulate a dataset with this characteristic and test the performance. We generate time series using sine and cosine waves as two classes (shown in Fig.~\ref{fig:sine_cosine_dataset2_sim}). Class 1 is generated by shifting a sine wave by $\pi$ and adding Gaussian noise with $\sigma=0.1$ and $\mu=0$. The second class is generated in the same fashion but a cosine wave is used, shifted by $\pi/3$, and Gaussian noise with $\sigma=0.2$ and $\mu=0$ is added. Time series in both classes have the same length of 100. A 20\%-80\% train-test split is used and the two classes are evenly split between train and test sets. For this experiment, DTW is optimal with $w=3$ and MDD is optimal with $\mathcal{E}=\{1,2,4\}$, with $\eta=0.3$. Looking at the performance, DTW achieves an accuracy of 0.47 on the train set and 0.60 on the test set. MDD, on the other hand, achieves an accuracy of 0.82 and 0.77 on the train and test sets, respectively.

        Another scenario when MDD may be a better choice concerns the separability of classes, particularly when the traces of time series are accompanied by a significant degree of noise. Noise can naturally decrease both the within-class and between-class separability. If this results in rendering the peaks and valleys non-discriminatory while still leaving differences to be recognized, DTW may not perform as effectively as expected. In contrast, the multi-scale nature of MDD may be able to look beyond the noise. An example of this can be seen in the SonyAIBORobotSurface1 dataset, where DTW's approach of matching peaks and dips fails to capture the differences between the two classes. In this dataset DTW only misclassified class 2 in one instance but misclassifies class 1 in more than 50\% of cases. This results in a precision of 0.99, 0.59, and a recall of 0.47, 1.0 for classes 1 and 2, respectively.
        
        \begin{figure}
            \centering
            \includegraphics[width=0.75\linewidth]{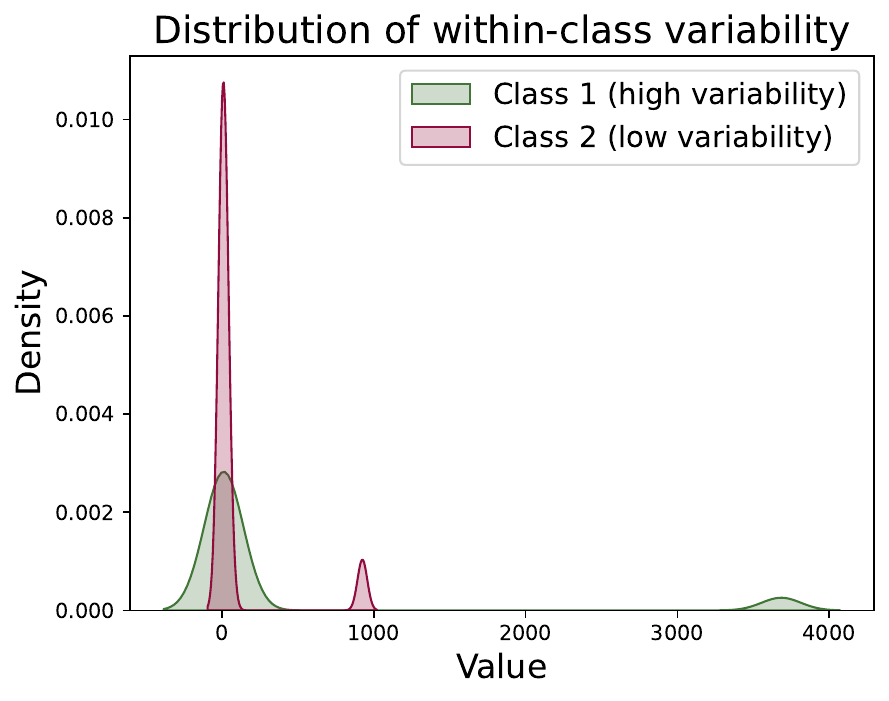}
            \caption{The distribution of $D_c$ (see Eq.~\ref{eq:within_class_sim}) on 10 datasets of UCR with visible difference in the within-class variability.}
            \label{fig:within_class_sim_10_dataset}
        \end{figure}

        \begin{figure}[h]
            \centering
            \includegraphics[width=0.7\linewidth]{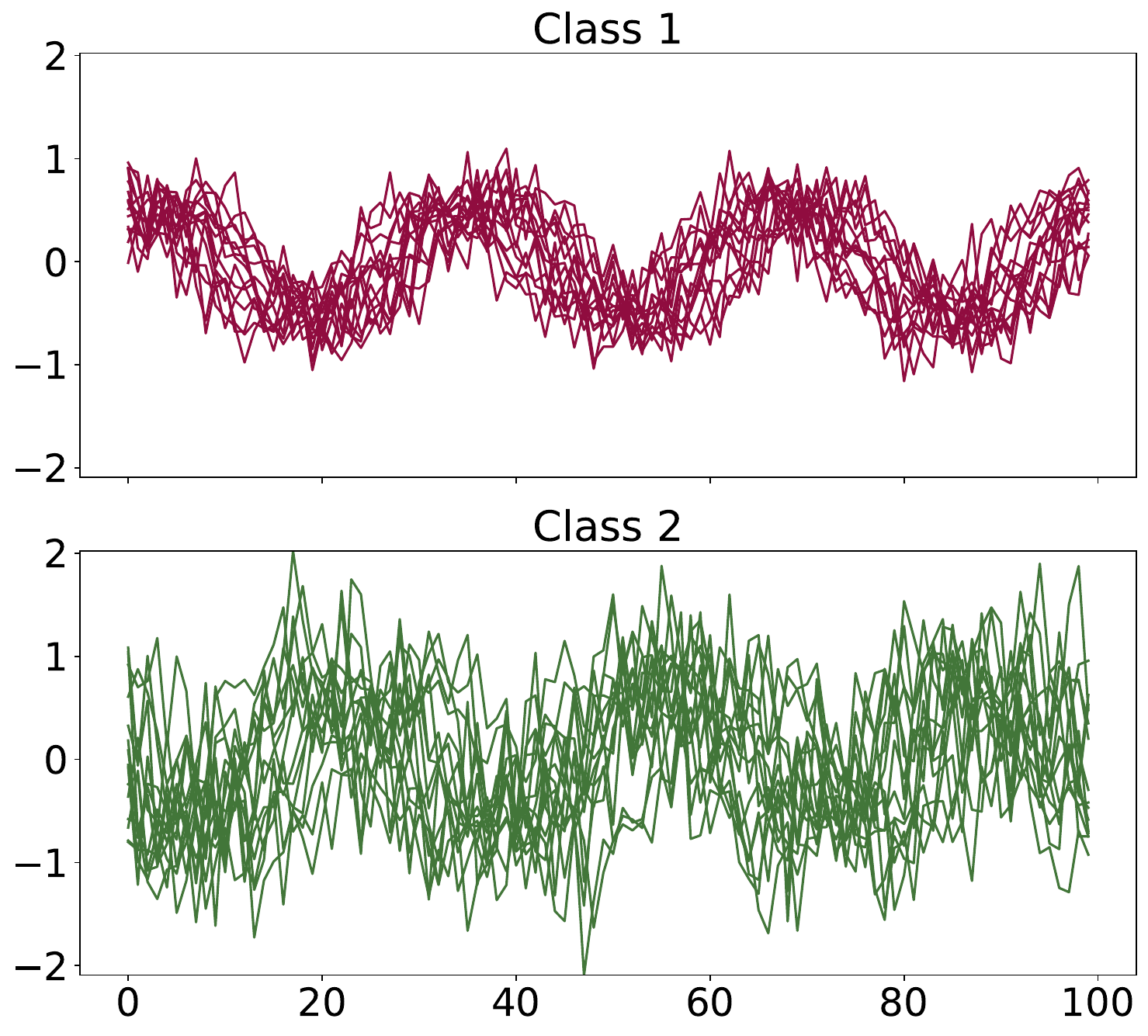}
            \caption{Examples of time series from a simulated data. Class 1 is generated with lower within-class variability while in general the classification of the classes using point-mapping strategy may not be effective. 1NN with DTW achieves accuracy 0.91 ($w=5$), 1NN with MDD achieves the accuracy of 0.97 ($\mathcal{E} = \{1, 2, 4\}$).}
            \label{fig:sine_cosine_var_dataset}
        \end{figure}

        To test this hypothesis, we compare the distribution of within-class variability of time series in several UCR datasets which visually exhibit the described characteristics. These datasets are FreezerRegularTrain (classes 1\&2), FreezerSmallTrain (1,2) , GunPointOldVersusYoung (1\&2), InsectEPGSmallTrain (1\&2), SonyAIBORobotSurface1 (1\&2), SwedishLeaf (2\&12  and 4\&10), ToeSegmentation1 (0\&1), Symbols (1\&2 and 4\&5), Worms (1\&3), DistalPhalanxOutlineCorrect (0\&1). We measure the within-class variability using Eq.~\ref{eq:within_class_sim} where $x_i$ and $x_j$ are two time series from class $c$ consisting of $n$ instances. $D_c$ is an $n\times n$ matrix of Euclidean distances between all pairs of time series in class $c$. We group all classes with higher variability together. The distribution of $D_c$, for high-variable classes versus low-variable classes is visualized in Fig.~\ref{fig:within_class_sim_10_dataset}. The narrower shape of the distribution corresponding to the low-variability classes shows that what we visually identified indeed reflects the within-class variability.

        \begin{equation}\label{eq:within_class_sim}
             D_{c} = \frac{2}{n (n-1)}\sum_{i < j}{EuD(x_i, x_j)}    
        \end{equation}
        
        To further test our hypothesis we simulate a dataset of sine and cosine waves in a similar fashion as before with the following configuration. Class 1 is generated by randomly shifting a sine wave by $2/3\pi$, adding Gaussian noise with $\sigma=0.2$ and $\mu=0$, and scaling down the amplitude by 0.6. The second class is a cosine wave randomly shifted by $\pi$, and Gaussian noise with $\sigma=0.4$ and $\mu=0$ is added, and scaling down the amplitude by 0.8. The resulting time series illustrated in Fig. \ref{fig:sine_cosine_var_dataset} shows high variability in class 2 and low variability in class 1. Running 1NN on this dataset, DTW achieves the accuracy score of 0.91 on the test set (with optimal $w = 5$), but MDD outperforms it on the same test set with the accuracy of 0.97 (with optimal $\mathcal{E} = \{1, 2, 4\}$).

    \subsection{A Real World Problem}
        As mentioned earlier, MDD is designed to fill a gap in time series similarity investigations: unlike most similarity measures used for time series, MDD does not rely on matching the points. In real-world problems, there are many cases in which matching-based definitions of similarity may not be appropriate. To demonstrate the effectiveness of MDD beyond the UCR archive, we use the SWAN-SF dataset, introduced in \ref{sec:data}.

        To ensure a fair evaluation, we select Partition 1 for training and Partition 2 for testing and preprocess as described in Sec.~\ref{sec:data}. Additionally, we apply $z$-Score standardization to time series in both partitions, utilizing only statistics from Partition 1.

        Our experimental setting is similar to those we carried out on the UCR archive, as discussed in Sec.~\ref{sec:mdd_dtw}. To compare the results, however, instead of the Texas Sharpshooter plot, we use two performance measures F1 Score, and  \textit{True Skill Statistic} (TSS) \cite{hanssen1965relationship} which is a domain-specific metric for space-weather forecasting applications. There is another domain-specific metric, namely \textit{Heidke Skill Score} (HSS) \cite{balch2008updated}, but since our dataset is balanced, it provides no extra information. TSS measures the difference between the probability of detection (i.e., \textit{true-positive-rate}), $\tfrac{tp}{p}$, and the probability of false alarm (i.e., \textit{false-positive-rate}), as shown in Eq.~\ref{eq:tss}.

        \begin{equation}\label{eq:tss}
            \textrm{TSS}=\frac{tp}{p} - \frac{fp}{n}
        \end{equation}

        To optimize DTW, we use the Sakoe-Chiba band, because it is the most commonly used warping window constraint \cite{keogh2002exact}. To find the optimal warping window, we trained 1NN for every possible window size, from 0 to 60 (i.e., 100\% of the time series length, equivalent to no-constraint).
        We determined $w = 22$ to be the optimal window size for this dataset with an F1 Score of 0.74. It should be noted that although we found 22 to be optimal, in general we observed that changing the warping window had little impact on the performance of DTW on SWAN-SF. All window sizes achieved F1 Scores within the narrow range of 0.735 to 0.741. To optimize MDD, we followed the strategy described in \ref{sec:mdd_dtw}. With $\eta=0.3$, we limited our search space to subsets of $\{1,2,4,8,16\}$. In cases where we observed ties in performance, we selected larger $\mathcal{E}$ because it would allow us to take advantage of MDD's multiscale property. With this approach, we determined $\mathcal{E} = \{1, 2, 4\}$ to be optimal, with an F1 Score of 0.80.

\begin{figure}
            \centering
            \includegraphics[width=0.8\linewidth]{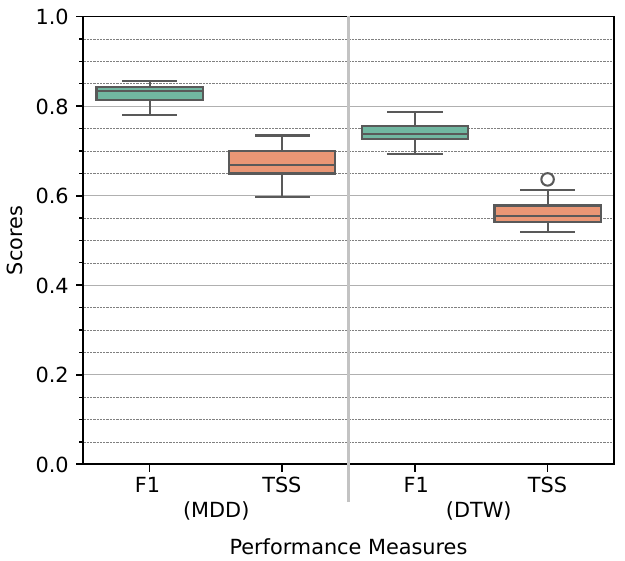}
            \caption{The boxplot shows 1NN with MDD outperforms 1NN with DTW on the SWAN-SF dataset.}
            \label{fig:SWANSF_F1_Scores}
        \end{figure}

        The results of our experiment is shown in Fig.~\ref{fig:SWANSF_F1_Scores}: 1NN coupled with MDD achieves a significantly higher average F1 Score of 0.83 ($\pm0.02$), to be compared with 0.74 ($\pm 0.03$). TSS for MDD reaches 0.67 ($\pm0.02$), and for DTW reaches only 0.63 ($\pm0.03$). Quantitatively, MDD is a better performance measure for SWAN-SF. Examining the time series, we can attribute this to the fact that the point-matching strategy does not seem to be an effective approach for measuring time series similarity in this dataset.

\section{Conclusion and Future Work}
In this paper, we examined a similarity metric for time series, namely Multiscale Dubuc Distance (MDD), that focuses on microscopic and macroscopic features of time series, simultaneously, without matching points. Comparing MDD's performance with DTW on 95 datasets from the UCR Time Series Classification Archive, we concluded that MDD's and DTW's wins are equally significant and not marginal. We discussed the overlap between the two measures as well as their complementary roles. Furthermore, we tested MDD's efficacy in time series classification on a real-world problem in solar physics in which point matching is not effective, and presented our results.
In this work we focused on using this metric for time series classification, but one extension of this work is utilizing MDD in indexing algorithms for content-retrieval databases. Further, we would like to investigate the effectiveness of MDD on other domains, and also on the operational flare-forecasting systems.
\section*{Acknowledgment}
This material is based upon work supported by the National Science Foundation under Grant No. 2209912, 2433781, and 2511630 directorate for Computer and Information Science and Engineering (CSE), and Office of Advanced Cyberinfrastructure (OAC). 

The authors would like to thank Prof. Eamonn Keogh and all people who contributed to the UCR Time Series Classification Archive.

\bibliographystyle{splncs04}
\raggedright
\bibliography{main}

\end{document}